%% file: main.tex
\documentclass[journal,onecolumn,12pt]{IEEEtran}
\usepackage{nopageno}
\usepackage{tabularx}
\usepackage{color,soul}
\usepackage{amsfonts} 
\usepackage{wrapfig}
\usepackage{longtable}
\usepackage{booktabs}
\usepackage{algpseudocodex}
\usepackage{algorithm2e}
\RestyleAlgo{ruled}
\usepackage{todonotes}

\newcommand{\measurement}{\mathbf{y}}
\newcommand{\signal}{\mathbf{x}}

\algrenewcommand\algorithmicrequire{\textbf{Input:}}
\algrenewcommand\algorithmicensure{\textbf{Output:}}

\usepackage{chngcntr}
\usepackage{cite}

\ifCLASSINFOpdf
  \graphicspath{{/figures/}{../pdf/}{../jpeg/}}
  \DeclareGraphicsExtensions{.pdf,.jpeg,.png}
\else
   \usepackage[dvips]{graphicx}
  \graphicspath{{figures/}{../pdf/}{../jpeg/}}
  \DeclareGraphicsExtensions{.eps,.png}
\fi
\usepackage[cmex10]{amsmath}
\begin{document}

\title{Accelerated Full Waveform Inversion by Deep Compressed Learning}

\renewcommand{\thefootnote}{\fnsymbol{footnote}} 

\author{Maayan Gelboim, Amir Adler, and Mauricio Araya-Polo\vspace{-0.7em}}

\maketitle

\thispagestyle{empty}

\begin{abstract}
We propose and test a method to reduce the dimensionality of Full Waveform Inversion (FWI) inputs as computational cost mitigation approach. Given modern seismic acquisition systems, the data (as input for FWI) required for an industrial-strength case is in the teraflop level of storage, therefore solving complex subsurface cases or exploring multiple scenarios with FWI become prohibitive.
The proposed method utilizes a deep neural network with a binarized sensing layer that learns 
 by \textit{compressed learning} a succinct but consequential seismic acquisition layout from a large corpus of subsurface models. Thus, given a large seismic data set to invert, the trained network selects a smaller subset of the data, then by using \textit{representation learning}, an autoencoder computes latent representations of the data, followed by K-means clustering of the latent representations to further select the most relevant data  for FWI. Effectively, this approach can be seen as a hierarchical selection. The proposed approach consistently outperforms random data sampling, even when utilizing only 10\% of the data for 2D FWI, these results pave the way to accelerating FWI in large scale 3D inversion. \end{abstract}
\section{Introduction}
Seismic inversion is a fundamental tool in geophysical analysis, involving the estimation of subsurface properties such as velocity, density and impedance   from recorded seismic data. Full Waveform Inversion (FWI) \cite{virieux2009} is a widely-used seismic inversion tool, and it has become the industry standard for model building/reconstruction in subsurface characterization workflows. It is utilized for hydrocarbon exploration, $CO_2$ sequestration and shallow hazard assessment, among others.
FWI has all the challenges of an inversion process: non-linear, non-convex, ill-posed, and it is computationally 
demanding\cite{virieux2009}, in particular for 3D inversion and when more complete physics beyond the acoustic approximation are targeted (i.e. elastic, viscoelastic). 

Modern seismic acquisition configurations with large number of seismic sources (i.e. shots) significantly increase the computational burden and complicate data manipulation. Several research directions aim to optimize seismic configurations and subsequent data manipulation to address these challenges. For example, using diffusion models for reconstruction and importance weighting of sources \cite{10733834}, or creating a sampling pattern based upon the changing 
complexity of the sampling area\cite{xu2024active}. 
 Some studies such as \cite{titova2023achieving} have explored the integration of Compressed Sensing (CS) based methods, where for instance,  random sampling of sources is performed, then other sources are added to the sampling space, controlling the maximum distance between sources as a parameter. In \cite{mark2017} a greedy algorithm is conditioned by computing illumination maps from virtual sources towards selecting an optimal subset of the data. A combined deep learning (DL) network and a straight-through estimator (STE) \cite{hubara2016binarized} to limit the number of sources and receivers when sampling shot gathers was suggested in \cite{10339330}. Thus, balancing between limiting the number of sources, but still optimize the selection of sources for the reconstruction of shot gathers. 
A combination of generative adversarial network (GAN) and CS was proposed in  \cite{8784224} that introduced shot gather reconstruction based on conditional-GAN combined with a repeated creation of sampling schemes to optimize the placement of sources near surface obstacles. In the same context, Jiang et al. \cite{10701294} utilized GAN to create a transformed domain that connects a CS-based sampling of the seismic data to the original, non-sampled data. Nevertheless, all the mentioned works focus on reconstructing of seismic data rather than inversion.

In previous work, we utilized Compressed Learning (CL) \cite{calderbank2009compressed}, which is a framework for machine learning in the CS domain,  
for DL-based 3D seismic inversion   \cite{Gelboim2023}. In this paper we leverage this approach, and augment it with \textit{representation learning} to reduce the FWI computational load, by choosing only a small subset of the available seismic data, this step is essential in realistic scenarios where very large volumes of seismic data, on the order of Terabytes, are typically utilized to reconstruct subsurface models. 

The contributions of this paper are two-fold: (i) we present a deep learning workflow that first learns candidate compressed sensing layouts, then selects online the best layout for specific data to invert by FWI; and (ii) we introduce the first utilization of \textit{representation learning} \cite{fard2020deep} for feature extraction and clustering \cite{lafabregue2022end} of seismic data in learned latent space for inversion. The rest of this paper is organized as follows: section II presents theoretical background on FWI and the main ingredients of the proposed solution: compressed learning and sensing, and representation learning. Section III presents the two-stage solution, including deep compressed learning and shot gathers clustering by representation learning. Performance evaluation is presented in section IV, and conclusions are discussed in section V. 

\section{Theoretical Background}
FWI is a seismic imaging technique that uses the full seismic wavefield to create high-resolution subsurface models. It works by iteratively adjusting a subsurface model $\hat{\mathbf{m}}$ until synthetic data generated from the forward operator $\textbf{F}(\hat{\mathbf{m}})$ matches the observed seismic data $\textbf{d}^{obs}$.
FWI minimizes the following loss function: 
\begin{equation}
J(\hat{\mathbf{m}})=\sum_{i=1}^{N_s}||\textbf{F}_i(\hat{\mathbf{m}}) - \textbf{d}^{obs}_i ||_2^2,\label{eq:fwi}
\end{equation}
where $N_s$ is the number of shots (i.e. seismic sources). Therefore, the complexity of minimizing $J(\hat{\mathbf{m}})$ is linear in $N_s$, which motivates FWI running time reduction by selecting only a small subset of the shots that are the most relevant for the inversion process.  
$\textbf{F}$ is a forward operator, which numerically solves seismic waves  propagation through the mechanical medium ($\hat{\mathbf{m}}$). 
In this work, we simulated seismic waves using the acoustic approximation \cite{tarantola_inversion_1984}, represented by the following wave equation: 
\begin{equation}
    \frac{\partial^2 \mathbf{u}}{\partial t^2} - \mathbf{V}\nabla^2 \mathbf{u} = \mathbf{f},
    \label{eq:acoustic_wave_eq}
\end{equation}
where $\mathbf{u}=\mathbf{u}(x,z,t)$ is the seismic wave displacement, $\mathbf{V}$ is the P-wave velocity model and $\mathbf{f}$ is the perturbation source (i.e. \textit{shot}) function. 
The relationship between the forward operator and the wave displacement can be described as:
$    \textbf{F}_i(\hat{\mathbf{m}}) = \mathcal{R}\mathbf{u}$
where $\mathcal{R}$ is a detection operator, responsible for obtaining the values calculated by the propagation simulation. 

To reduce the number of utilized shots we designed a \textit{compressed learning} \cite{calderbank2009compressed} deep  network, detailed in section III, that performed \textit{compressed sensing} of the shots.  
\textit{Compressed Sensing} (CS) enables reconstruction of a signal from a small number of linear projections (i.e. measurements) measurements, under certain assumptions as detailed in the following. Given a signal $\signal \in \mathbf{R}^N$ an $M \times N$ sensing matrix $\Phi$  (such that $M \ll N$) and a measurements vector $\measurement = \Phi \signal$, the goal of CS is to recover the signal from its measurements. According to CS theory \cite{1614066}, signals that have a sparse representation in the domain of some linear transform can be exactly recovered with high probability from their measurements. While CS was originally developed for general sensing matrices, it was  extended \cite{8052513} to  binary sensing matrices. \textit{Compressed Learning} (CL) is the extension of CS to solve machine learning problems in the CS domain. CL was introduced in \cite{calderbank2009compressed}, which proved that direct inference from the compressive measurements $\textbf{y} = \Phi \textbf{x}$ is possible  with high accuracy in Support Vector Machines. CL extensions to deep learning were introduced by \cite{adlercompressed,zisselman2018compressed}. 

In this work we employed for the first time \textit{representation learning} (RL)\cite{6472238} to analyze seismic traces in latent space. RL is utilized for automatic discovery of useful features from raw data, without relying on manual feature engineering. For RL we utilized an autoencoder, composed by a cascade of encoder and decoder sub-networks. The encoder $f_E$ maps a seismic signal $\textbf{x}$ to a reduced dimension latent representation $\textbf{z}$, namely $\textbf{z} = f_E(\textbf{x})$. The decoder reconstructs the signal $\textbf{x}$ from the latent representation, namely $\hat{\textbf{x}} = f_D(\textbf{z})$. The latent representation $\textbf{x}$ retains essential information, enabling tasks such as denoising, compression, anomaly detection and inference, among others, to be performed \cite{fard2020deep, lafabregue2022end}.

\begin{figure*}[ht!]
    \centering
\includegraphics[scale=0.32]{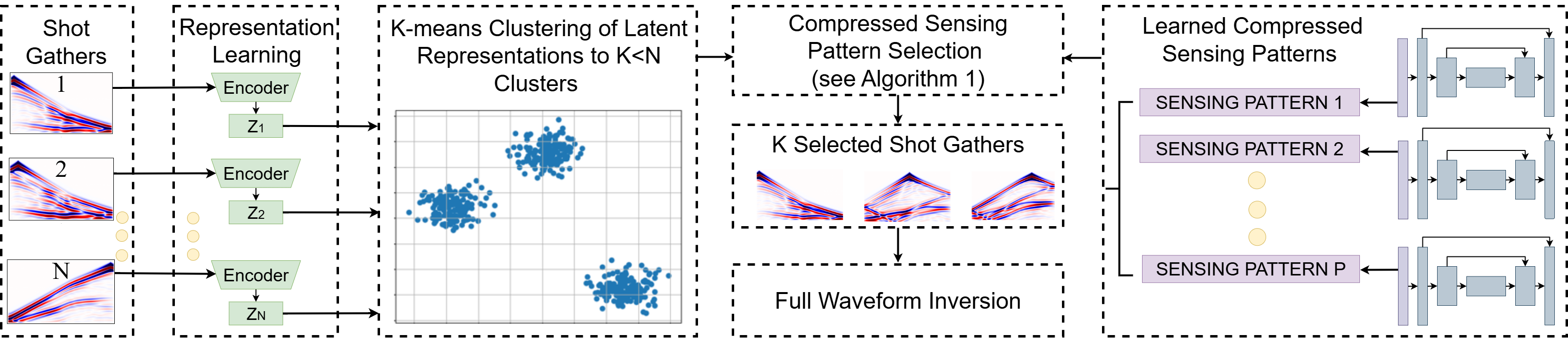} 
\vskip -9pt
\caption{The proposed DCL-RL solution evaluates sensing patterns, obtained from trained deep compressed learning models. Using the latent representations of the shot gathers, we assign scores to each learned sensing pattern to identify the most informative subset of shot gathers for reconstructing a velocity model.}
     \label{fig:Process}
\vspace{-1em}
\end{figure*}

\section{Deep Compressed Leaning in FWI}
We designed a 2D Deep CL (DCL) architecture (Table~\ref{tab:U-NET}) that jointly optimizes shots selection and the inversion of the selected shots. A sensing layer for shot selection was designed using a \textit{straight-through estimator} \cite{hubara2016binarized} to create learnable binary weights for all available shots (i.e. '1' indicates a selected shot and '0' non-selected shot). This layer utilizes a binarization function, $\phi_b(w) = \textbf{1}(w>0)$, returning '1' if $w>0$ and '0' otherwise, where $w \in \mathbb{R}$ is a learnable weight. The binarization function's derivative is discontinuous and has an indefinite value, making it unsuitable for back-propagation. As a result, we employ the binarization function for the forward-pass but substitute the hard-sigmoid function \cite{hubara2016binarized} for the backward-pass.
This architecture was trained by minimizing the following mixed loss function:
\begin{equation}
 \mathcal{L}(\mathbf{m_{i}},\hat{\mathbf{m_{i}}},R,\hat{R}) = \ \text{MAE}(\mathbf{m_{i}},\hat{\mathbf{m_{i}}}) + \mu(R-\hat{R})^2,
 \label{loss_eq}
\end{equation}
where MAE is mean absolute error, $\mathbf{m_{i}},\hat{\mathbf{m_{i}}}$ represent the ground-truth and predicted velocity models, respectively, $R,\hat{R}$ are the target and learned sensing rates, respectively, and $\mu>0$ controls the trade-off between the two misfit terms.  The learned sensing rate is defined as $\hat{R}=\frac{1}{N_s}\sum_{i=1}^{N_s}{\phi_b(i)}$, where 
$\phi_b(i)$ is the \textit{i}-th  binarized coefficient of the sensing layer. 
The binarized coefficient's gradient is estimated using a STE. Training the DCL model yields two outputs: a set of selected shot gathers (i.e. learned sensing pattern) and a reconstructed velocity model, not used in this work. 

We observed that training the DCL multiple times resulted in different sensing patterns for velocity model reconstruction with comparable high quality (SSIM $> 0.9$). This variability, namely having several equally good sensing patterns, is reasonable and was also reported by \cite{titova2023achieving} in the context of CS for shot gather reconstruction. 
Therefore, for all the velocity models in the training data, we can learn multiple sensing patterns of shot gathers, by multiple training of the DCL network. However, applying each one of the learned patterns on new unseen data (i.e. selecting different subsets of shots), will not necessarily provide equally good results, and requires identifying the best sensing pattern for the specific data to invert by an additional selection step.

Let  $\{\theta_1, \theta_2, ..., \theta_P\}$ denote a set of $P$ vectors, where each $\theta_i \in \{1,2,...,N\}^K$ represents a sensing pattern for selecting $K$ shot gathers from a total of $N$ sources, aimed at achieving high-quality reconstruction of the velocity model $\hat{\textbf{m}}$. Let $S =\{X_1, X_2, ..., X_N\}$ denote a set of $N$ shot gathers, here each $X_i \in \mathbb{R}^{R\times T}$, where $R,T$ are the number of receivers and time-samples, respectively.

\textbf{Representation learning:} we employed a convolutional auto-encoder (CAE) for mapping shot gathers $X_i$ to compact latent representations (Table \ref{tab:AE}). The encoder sub-network $E$ projects the shot gather into a low-dimensional latent representation $\textbf{z}_i = E(X_i)$, where $\textbf{z}_i\in\mathbb{R}^{F\times H\times W}$ and  $F\times H \times W < R\times T$. The decoder sub-network (including the bottleneck) $D$, reconstructs the CAE input as $\hat{X_i} = D(\textbf{z}_i)$. 


Applying K-means on the latent representations to create $K$ clusters is denoted by $f_{Kmeans}(\textbf{z}_1, ..., \textbf{z}_N)$, which results in $N$ labels (one per shot): $L =\{\hat{y}_{(\textbf{z}_1)},..,\hat{y}_{(\textbf{z}_N)}\}$ where $\hat{y}_{(\textbf{z}_i)} \in \{1,2,...,K\}$. Let us denote a set $S_j$ as a subset of shot gathers that belong to class $j \in \{1,..,K\}$. The subset \(L_{\theta_i}\) includes only the labels associated with the shot gathers specified by \(\theta_i\), defined by:
\begin{equation}
L_{\theta_i} = \left\{ \hat{y}_{(z_j)} \in L \;\middle|\; j \in \theta_i \right\}.
\end{equation}
The underlying assumption is that a highly informative sensing pattern would include shot gathers from as many clusters as possible. To quantify this, we assign a \textit{diversity }score \(s_{1\theta_i}\) to each recommendation vector \(\theta_i\), calculated as the number of unique elements in \(L_{\theta_i}\): 
$s_{1\theta_i} =  \left|\text{unique}(L_{\theta_i})\right|,$
and selected the sensing pattern vector with the highest diversity score. In cases where multiple vectors achieve the same maximum value of \(s_{1\theta_i}\), we introduce a \textit{distance} score \(s_{2\theta_i}\), defined as the sum of pairwise Euclidean distances between the latent representations of the shot gathers within \(\theta_i\): 
\begin{equation}
s_{2\theta_i} = \sum_{n\in\theta_i}\sum_{m\in\theta_i} \| \textbf{z}_n - \textbf{z}_m\|_2.
\end{equation}
The underlying assumption is that higher average distance between shot gathers in the latent space, namely a higher \(s_{2\theta_i}\) score,  indicates that more unique information is carried by these shots, and therefore more relevant for inversion. 
The complete solution is summarized in Fig. 1 and Algorithm 1.\\


\setcounter{algocf}{0}
\SetKwComment{Comment}{/* }{ */}
\newcounter{stepnum}
\newcommand{\step}[1]{\refstepcounter{stepnum}\textbf{\thestepnum:} #1}

\begin{center}

\begin{minipage}{0.6\linewidth}
\begin{algorithm}[H]
\caption{Shots Sensing Pattern Selection}

\KwIn{$\{\theta_1, \dots, \theta_P\}$: A set of $P$ sensing patterns\\
      $\{X_1, X_2, \dots, X_N\}$: A set of $N$ shot gathers\\
      $E$: Encoder from a trained shots Autoencoder\\
      $K$: Target number of selected shot gathers ($K<N)$}
\KwOut{$\theta_{i^*}$: Selected compressed sensing pattern}
\BlankLine
\step{Encode all shot gathers to latent representations:}\\
$\{\textbf{z}_1, \dots, \textbf{z}_N\} \gets E(X_1, \dots, X_N)$\;
\BlankLine
\step{Cluster the latent representations to K clusters:} \\
$L \gets \{\hat{y}_{(\textbf{z}_1)}, \dots, \hat{y}_{(\textbf{z}_N)}\} = f_{Kmeans}(\textbf{z}_1, \dots, \textbf{z}_N)$\; 
\BlankLine
\step{Compute \textit{diversity} score by cluster representation:} \\
\For{$i \gets 1$ \KwTo $P$}{
    $s_{1\theta_i} \gets \left|\text{unique}\left(\{\hat{y}_{(z_j)} \in L \mid j \in \theta_i\}\right)\right|$\;
}
$i^* \gets \arg\max_{i \in \{1, \dots, P\}} s_{\theta_i}$\;

\BlankLine
\step{Compute \textit{distance} score by Euclidean distances:} \\
$s_{2\theta_{i^*}}  \gets \sum_{n\in\theta_{i^*}}\sum_{l\in\theta_{i^*}} \| \textbf{z}_n - \textbf{z}_l\|_2$\;

\For{$q \gets 1$ \KwTo $P$}{
\If{$s_{1\theta_{i^*}} = s_{1\theta_q}$ \textbf{and} $ i^*\neq q$ }
{
    $s_{2\theta_q}  \gets \sum_{n\in\theta_q}\sum_{m\in\theta_q} \| \textbf{z}_n - \textbf{z}_m\|_2$\;
    \If{$s_{2\theta_q} > s_{2\theta_{i^*}}$ }
    {
    $i^* \gets q$
    }
}
}
\Return{$\theta_{i^*}$}\;
\label{alg: best rec}

\end{algorithm}
\end{minipage}
\end{center}
\vspace{-1em}

\subsection{Dataset Preparation} 
We created a dataset of 5,800 layered 2D velocity models with 5-8 layers using GemPy \cite{gempy19}, with velocity ranging between 2 to 4.5 km/s. The dataset was split into the following disjoint sets; deep learning models (DCL and autoencoder) training and validation 5500 and 250 models, respectively, and 50 testing models for FWI evaluation. Each velocity model represented an area of 2 km $\times$ 1 km (width $\times$ depth) discretized to a grid of 288 $\times$ 144 points. The spacing between grid points was set at 6.99 m. Receivers were placed at every other grid point along the lateral dimension. Each shot gather was recorded for a period of one second. For each velocity model, 20 shot gathers were generated, with shots positioned uniformly  along the 2km lateral dimension. 


\subsection{Deep Learning Architectures and Training}
Tables~\ref{tab:U-NET}
 and~\ref{tab:AE} present the two deep learning architectures used in our experiments. Table~\ref{tab:U-NET} describes our proposed DCL architecture, where all convolution layers include ReLU activations. IN* denotes instance normalization layer, and Enc1* denotes an encoder layer without max pooling. 
For the STE, we initialized the model randomly while ensuring that the number of initially selected shot gathers matched the target value. This design enables monitoring of how far the STE deviated from its initial configuration during training. The initial weight for a binary '1' was set to 0.25, and for a binary '0' to -0.25.
Table~\ref{tab:AE} presents the autoencoder architecture used to encode shot gathers (output of Enc5) for K-means clustering. 
Prior to training, all shot gathers were normalized to  [-1,1] for consistent input scaling. 
ADAM optimizer was used for training all deep learning models, experiments were conducted on NVIDIA GraceHopper GH200 GPUs.

\begin{table}[ht!]
\caption{Deep Compressed Learning Architecture}
\footnotesize
\centering
\begin{tabular}{|c|c|c|c|}
\hline
 Block & Layer &  Unit &  Comments \\\hline
 Input & 0 &  20 shot gathers &  20x864x144 \\
      &  &  & grid points \\\hline
 Sensing & 1 & Binarized Sensing Layer &  20 learnable\\
 &  &  & parameters \\\hline
 Enc1 & 2 & Conv2D(32, (5x5)) & + IN* \\
 & 3 & Conv2D(32, (5x5)) & + IN* \\
 & 4 & MaxPool2D & + Dropout(0.2) \\\hline
 Enc2 & 5-7 & Enc1(64) & \\\hline
 Enc3 & 8-10 & Enc1(128) & \\\hline
 Enc4 & 11-13 & Enc1(256) & \\\hline
 Enc5 & 14-15 & Enc1*(512) & \\\hline
 Dec1 & 16 & ConvTrans2D(256, (2x2)) & + IN* \\
 & 17 & Conv2D(256, (2x2)) & + IN*  \\
 & 18 & Conv2D(256, (2x2)) & + IN* \\\hline
 Dec2 & 19-21 & Dec1(128) & \\\hline
 Dec3 & 22-24 & Dec1(64) & \\\hline
 Dec4 & 25-27 & Dec1(32) & \\\hline
 & 28 & Conv2D(1,(1x6)) & stride=(1x6)\\
 &   &  & dilation=(1x1)\\\hline
 output & 29 & Velocity Model & 288x144 grid points\\\hline

\end{tabular}
\small
\label{tab:U-NET}
\vspace{-5pt} 
\end{table}

\begin{table}[ht!]
\caption{Convolutional Autoencoder for\\ Shot Gathers Representation Learning}
\centering
\begin{tabular}{|c|c|c|}
\hline
 Block & Layer &  Unit \\\hline
 Input & 0 &  Shot gather  \\\hline
 Enc1 & 1 &  Conv2D(16,(3x3),ReLU)   \\
      & 2 & MaxPool2D  \\\hline
 Enc2 & 3-4 &  Enc1(8)   \\\hline
 Enc3 & 5-6 &  Enc1(8)   \\\hline
 Enc4 & 7-8 &  Enc1(8)   \\\hline
 Dec1 & 9 &  Conv2D(8,(3x3),ReLU)  \\
      & 10 & UpSampling2d  \\\hline
 Dec2 & 11-12 &  Dec1(8)   \\\hline
 Dec3 & 13-14 &  Dec1(8)   \\\hline
 Dec4 & 15-16 &  Dec1(16)   \\\hline
 Dec5 & 17 &  Conv2D(1,(3x3),Tanh)   \\\hline
 Output & 18 & Reconstructed Shot Gather  \\\hline  
\end{tabular}
\label{tab:AE}
\vspace{-8pt} 
\end{table}

\section{Performance Evaluation}
As a reference for all experiments, FWI experiments were conducted using the complete set of 20 shot gathers on the 50 test velocity models (not used during training or validation of the deep learning models). Following standard practice, we initialized the FWI executions with smoothed version of the ground-truth velocity model. We then repeated the FWI experiments on the same testset of velocity models, but using randomly selected subsets of 2, 3, 4 and 5 shot gathers. For each of these configurations, the shot gathers were randomly chosen prior starting FWI. To ensure statistical robustness, we repeated the random selection experiments three times for each velocity model. 

\begin{figure*}[ht!]
    \centering    \includegraphics[width=0.97\linewidth]{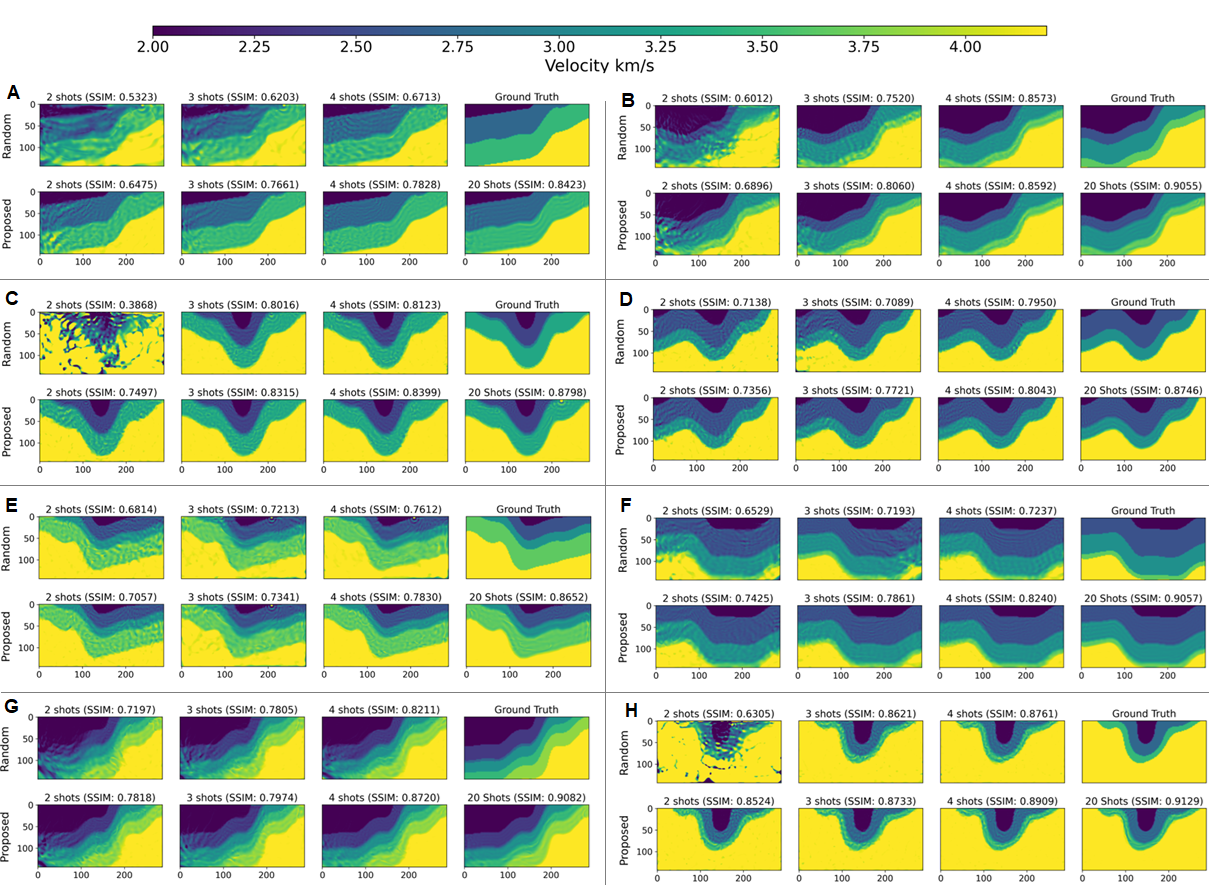}
    \vskip -10pt
    \caption{FWI results: per each velocity model (A-H) the ground truth and inversion using all (20) shot gathers are compared vs. inversion with 4 shots (20$\%$ sensing rate), 3 shots (15$\%$) and 2 shots (10$\%$), clearly indicating the advantage of the proposed DCL-RL approach over random sampling. \label{fig:examples}}     
\vspace{-0.4em}
\end{figure*}

All FWI experiments were undertaken following stopping conditions: (i) Early stopping after 10 iterations without decrease in the loss function $J(\mathbf{m})$; or (ii) Early stopping upon reaching $J(\mathbf{m}) \leq 0.05$, a threshold beyond which we observed no significant improvements in velocity model reconstruction; or (iii) a limit of 500 iterations.  All FWI experiments were performed on computing nodes supporting AMD Genoa-X multi-core. 
We trained three times each DCL architectures  (i.e., for 2, 3, 4 and 5 shot gathers).  
This resulted in a total of twelve distinct shot gather sensing layers (three per sensing rate). Since we used 50 velocity models for testing, this lead to 150 FWI runs per sensing rate. 
Next, FWI experiments were carried out using each DCL-learned sensing layer, with standard smoothed initial velocity model. In the final stage, we applied the proposed solution (DCL-RL) to select the shots for FWI, and evaluated inversion quality. Fig. \ref{fig:examples} presents examples of FWI results under the different strategies for shot selection, and Fig. 3 summarizes measured FWI running times for the different number of selected shots, clearly indicating the potential of accelerating FWI. 
Table \ref{fwi_res} summarizes the results of all experiments above. Inversion quality trends can be observed.  as follows: (i) as expecred, FWI utilizing a randomly selected subset of the available shots   produced less accurate models as compared to using all 20 shots. (ii) DCL-based shot selection vs. random selection, is sometimes better but not for all sensing rates. (iii) the proposed DCL-RL solution consistently outperforms random shot selection in terms of mean absolute error (MAE, lower is better), structural similarity (SSIM, higher is better) and Peak Signal-to-Noise Ratio (PSNR, higher is better), for all sensing rates. 
\vspace{-0.3em}
\section{Conclusions}

\begin{figure}[ht!]
    \centering
    \includegraphics[trim={0 0 0 0.72cm},clip,width=0.5\linewidth]{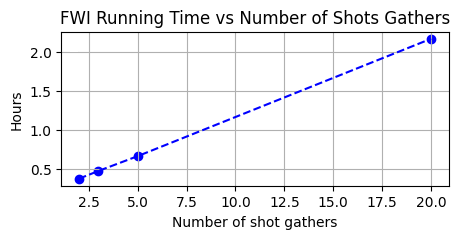}
    \vskip -12pt
    \caption{FWI running time on AMD Genoa-X multi-core vs. number of utilized shot gathers.}
    \label{fig:placeholder}
\end{figure}
We presented a novel solution for shot selection in FWI by utilizing auxiliary  compressed learning and representation learning deep neural networks. The proposed DCL-RL solution achieved consistent advantage in FWI model reconstruction quality as compared to random shot selection, which is often used in practice.
Our study also reveals that the advantage of DCL-RL increases for the lower sensing rates (e.g. almost 3dB PSNR advantage at 10\% sensing rate). These results pave the way for FWI acceleration in large scale surveys with very large numbers of shots gathers, especially (but not limited to) 3D surveys. 
Future work should evaluate this method on larger and more diverse 2D and 3D testsets.

\input{fwi_results_latest_captioned}
\section{ACKNOWLEDGMENTS}
The authors thank TotalEnergies EP R\&T US for support and permission to share this material.
\bibliographystyle{ieeetr}
\bibliography{biblio}
\end{document}

%% file: fwi_results_latest_captioned.tex
\begin{table}[ht]
\centering
\caption{FWI results  with different sensing rates and shot selection methods: DCL-RL (Deep CL + Representaion Learning + K-means), DCL (Deep CL), Random Selection. All results  averaged over 50 test velocity models.}
\begin{tabular}
{|l|c|c|c|c|c|}
\hline
 Shots (Rate) &     Selection &    MAE  &  SSIM  & PSNR [dB]\\\hline
     2  (10\%) & DCL-RL &       \textbf{0.086980} &  \textbf{0.69406} & \textbf{30.265}\\
     2 &           DCL &      0.128500 &  0.63556 & 27.590\\
     2 &        Random &      0.128720 &  0.64132 & 27.559\\
     \hline
     3 (15\%)& DCL-RL &       \textbf{0.06928} &  \textbf{0.75678} & \textbf{32.654}\\
     3 &           DCL &      0.0767 &  0.74249 & 32.141\\
     3 &        Random &      0.0911 &  0.70532 & 30.402\\
     \hline
     4 (20\%)& DCL-RL &       \textbf{0.06306} &  0.76140 & 32.862 \\
     4 &           DCL &      0.06387 &  \textbf{0.76695} & \textbf{33.021}\\
     4 &        Random &      0.07548 &  0.74934 & 31.925 \\
     \hline
     5 (25\%)& DCL-RL &       \textbf{0.05076} &  \textbf{0.79774} & \textbf{34.355}\\
     5 &           DCL &      0.06718 &  0.77216 & 33.004\\
     5 &        Random &      0.05776 &  0.78548 & 33.506\\
     \hline
    20 (100\%)&           All &       0.03306 &  0.86592 & 36.597\\
\hline
\end{tabular}
\label{fwi_res}
\end{table}